\documentclass[11pt]{article}

\usepackage{acl}
\usepackage{times}
\usepackage{latexsym}
\usepackage[utf8]{inputenc} 
\usepackage[T1]{fontenc}    
\usepackage{hyperref}       
\usepackage{url}            
\usepackage{booktabs}       
\usepackage{amsfonts}       
\usepackage{nicefrac}       
\usepackage{microtype}      
\usepackage{xcolor}         
\usepackage{graphicx}
\usepackage{multirow}
\usepackage{inconsolata}
\usepackage[most]{tcolorbox}
\usepackage{subcaption}
\usepackage{fontawesome5}

\newcommand{\CODEURL}{https://github.com/JigglypuffStitch/ScreenHaystack}
\newcommand{\HFURL}{https://huggingface.co/datasets/Charlotte188/ScreenHaystack} 

\newtcolorbox{promptbox}[1]{
    colback=gray!3,
    colframe=gray!45,
    boxrule=0.4pt,
    arc=1.5pt,
    left=5pt,
    right=5pt,
    top=5pt,
    bottom=5pt,
    title=#1,
    fonttitle=\bfseries\small,
    fontupper=\small,
    breakable
}

\newtcolorbox{findingbox}{
    enhanced,
    colback=black!2,
    colframe=black!8,
    boxrule=0.25pt,
    sharp corners,
    borderline west={1.2pt}{0pt}{black!55},
    left=7pt,
    right=7pt,
    top=6pt,
    bottom=6pt,
    before skip=0.8em,
    after skip=0.8em
}

\title{\textsc{ScreenHaystack}: Finding Blind Zones in GUI Grounding}

\author{
Chenyue Li,
Xiaoxiao Sun,
Yubo Deng,
Qinlin Zhao,
Serena Yeung-Levy,
Yuhui Zhang \\
Stanford University\\
\faGithub~\href{\CODEURL}{Code} \hspace{0.25em}
\faDatabase~\href{\HFURL}{Data}
}

\begin{document}

\maketitle

\begin{abstract}
We introduce \textsc{ScreenHaystack}, a dynamic needle-in-a-haystack benchmark for evaluating spatial reliability in GUI grounding. Instead of testing each target at a fixed position, \textsc{ScreenHaystack} systematically relocates controlled target icons across high-resolution GUI backgrounds and measures whether models can localize them consistently. Using this benchmark, we find that leading GUI grounding models, including Qwen3-VL, UI-TARS, GTA, and UI-Venus, exhibit \emph{blind zones}: spatial regions where grounding accuracy drops sharply despite fixed target appearance and instruction. These blind zones transfer to unseen ScreenSpot-Pro examples: targets inside blind zones are consistently harder to ground, with Qwen3-VL-8B dropping by 16.1 percentage points, and controlled relocation shows that moving targets into blind zones decreases accuracy while moving them out improves accuracy. We further show through controlled synthetic experiments that uneven spatial coverage in training data can induce such blind zones. Therefore, we propose a simple strategy, blind-zone-oriented augmentation, which adds supervision in blind zones and improves ScreenSpot-Pro accuracy over both original and randomly augmented Click-100k fine-tuning.

\end{abstract}

\section{Introduction}

GUI grounding is a core capability for computer-using agents: given a screenshot and an instruction, a model must localize the target UI element for actions such as clicking, typing, or scrolling. This capability underlies recent progress in GUI agents and multimodal computer-use systems~\citep{nguyen2025gui,cheng2024seeclick,qin2025ui,gelato2025,sun2026learning}. Existing grounding benchmarks such as ScreenSpot-Pro~\citep{li2025screenspot} have accelerated this progress by evaluating grounding on realistic interfaces.

\begin{figure}[t]
    \centering
    \includegraphics[width=\linewidth]{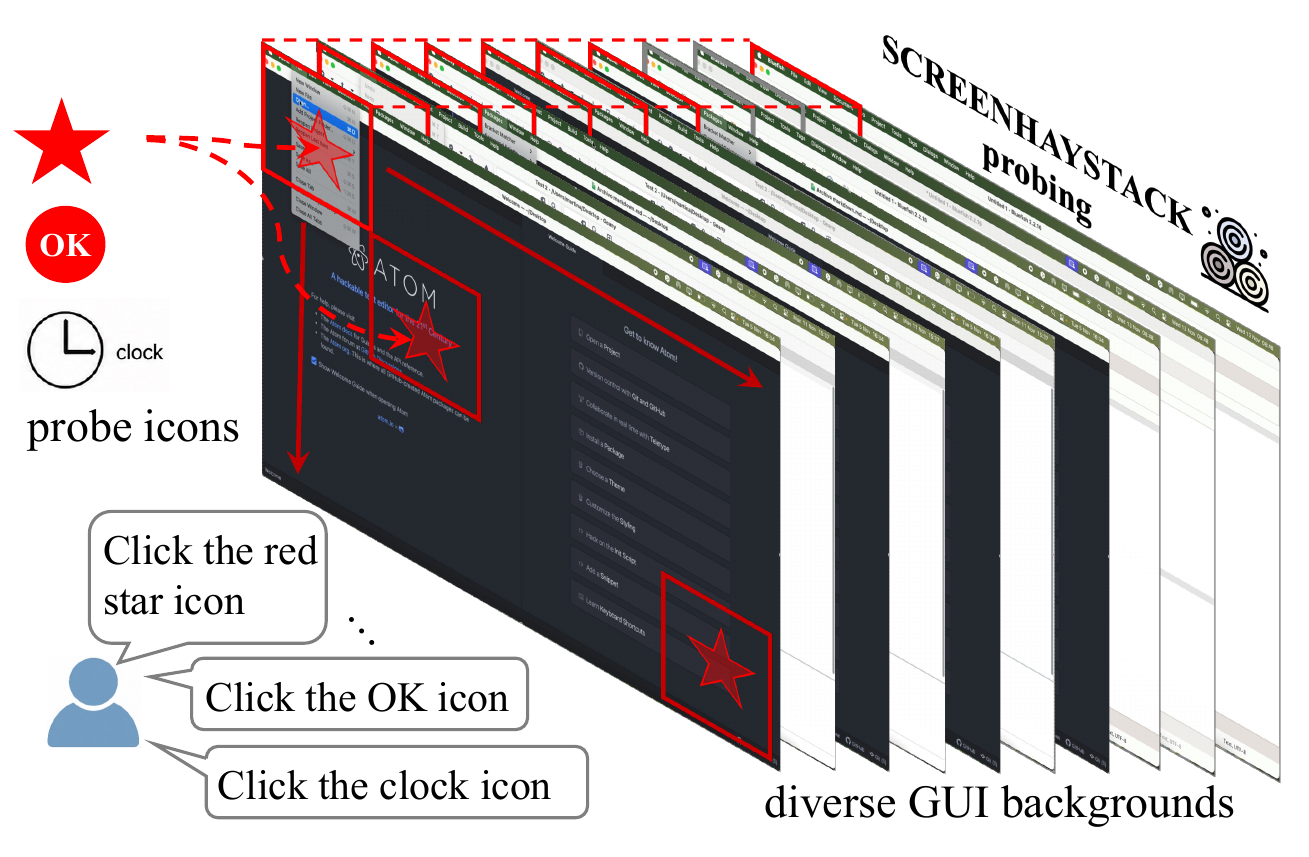}
      
  \caption{\textsc{ScreenHaystack probing.} 
Controlled probe icons are relocated across diverse GUI backgrounds to measure location-dependent grounding accuracy while reducing the influence of icon appearance and background content.}
    \label{fig:dataset}
\end{figure}

\begin{figure*}[t]
    \centering
    \includegraphics[width=1\linewidth]{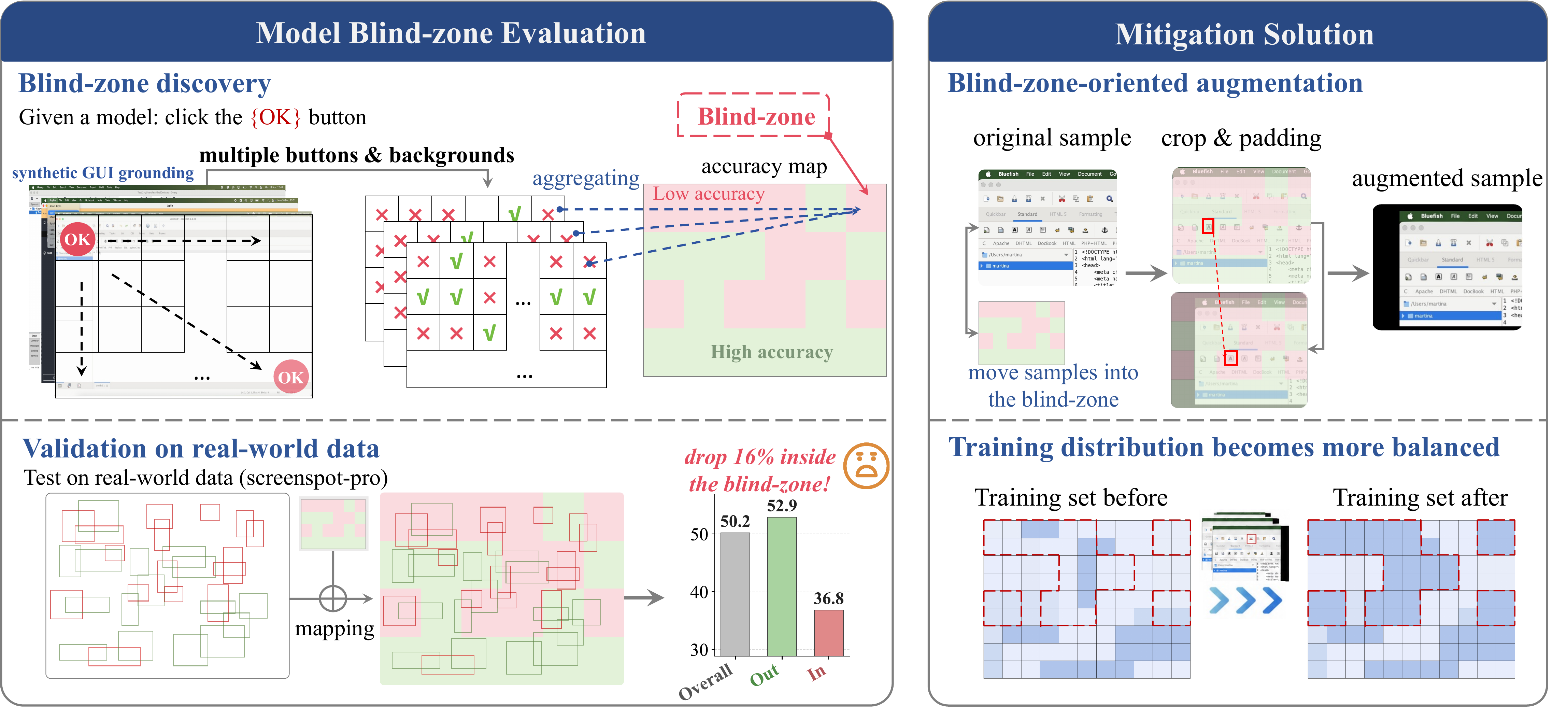}
    \caption{\textbf{Overview of \textsc{ScreenHaystack}}.
   \textbf{Left:} \textsc{ScreenHaystack} moves controlled target icons across diverse GUI backgrounds and aggregates model predictions into spatial accuracy maps to identify model-specific blind zones. These zones are validated on ScreenSpot-Pro, where targets inside them are harder to localize. \textbf{Right:} Blind-zone-oriented augmentation uses the identified blind zones to guide crop-and-padding transformations, moving training targets into weak regions to improve spatial coverage and grounding reliability.}
   
    \label{overview}
\end{figure*}

Yet most grounding benchmarks are static: each target appears at one fixed position, and evaluation reports a single score. This protocol is valuable for measuring general grounding ability, but it collapses spatial reliability into one score. In real interfaces, element locations are not fixed: windows resize, pages scroll, layouts reflow, and applications enter different states. A robust grounding model should therefore localize the same target reliably across the image plane, not only at the positions provided by static benchmarks. This distinction matters for deployed agents: a model with strong average accuracy can still be unsafe if its errors concentrate in screen regions where common controls may appear after resizing or scrolling.

To evaluate spatial reliability in GUI grounding, we introduce \textsc{ScreenHaystack}, a dynamic needle-in-a-haystack benchmark. \textsc{ScreenHaystack} inserts controlled target icons into diverse high-resolution GUI screenshots and systematically relocates them across a spatial grid (Figure~\ref{fig:dataset}). Because target identity and instruction are held fixed while location changes, the benchmark directly measures how grounding accuracy varies as a function of target position. Aggregating results across backgrounds and icons reveals model-specific hard regions, which we call \emph{blind zones}. Figure~\ref{overview} (left) shows the process of relocating controlled targets and aggregating predictions to identify blind zones for a given model.

Using \textsc{ScreenHaystack}, we evaluate several state-of-the-art GUI grounding models, including Qwen3-VL~\citep{bai2025qwen3}, UI-TARS~\citep{qin2025ui}, GTA~\citep{yang2026gta1}, and UI-Venus~\citep{gu2025ui}. Results indicate distinct blind zones where grounding accuracy significantly decreases, with failures not evenly distributed across the screen: each model has a structured, model-specific pattern of weak regions.

We further show that these blind zones are not artifacts of synthetic probing. When blind-zone masks are projected onto unseen ScreenSpot-Pro data, 
targets inside the masks are consistently harder to ground than targets outside them. As shown in Figure~\ref{overview} (bottom-left), Qwen3-VL-8B drops by 16.1 percentage points inside blind zones, and other models show similar drops of 7.1 to 14.3 points. To test the spatial effect, we relocate ScreenSpot-Pro targets across blind-zone boundaries. Moving targets into blind zones decreases accuracy by 17.6 points on average, while moving targets out of blind zones improves accuracy by 8.1 points on average. Within-region random-shift controls produce much smaller changes.

Finally, we study uneven spatial coverage in training data as one plausible cause of blind zones. Since most strong GUI grounding models are trained on private or mixed data, we use controlled synthetic fine-tuning experiments where target coverage is intentionally removed from selected regions. These experiments show that uneven spatial coverage alone can induce blind-zone-like failures, especially when supervision is missing jointly along both coordinate dimensions. Motivated by this finding, we propose blind-zone-oriented augmentation (Figure~\ref{overview} right), which adds training examples in weak spatial regions. On Click-100k fine-tuning, this strategy improves ScreenSpot-Pro accuracy to 61.47\%, outperforming both original fine-tuning and random augmentation.

Our contributions are threefold. First, we introduce \textsc{ScreenHaystack}, a dynamic benchmark that evaluates GUI grounding as a function of target location. Second, we identify blind zones in current GUI grounding models and show that these probe-discovered regions align with failure patterns on unseen ScreenSpot-Pro data and affect grounding accuracy under controlled relocation. Third, we show that uneven spatial coverage can induce blind zones and propose blind-zone-oriented augmentation as a simple mitigation strategy.

\begin{figure*}[t]
    \centering
    \includegraphics[width=\linewidth]{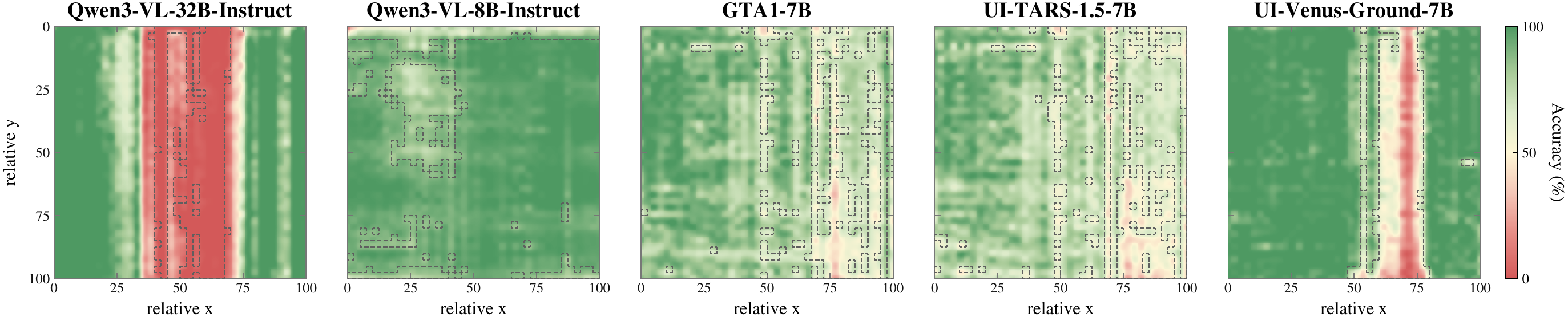}
\caption{\textbf{Model-specific blind-zone maps from \textsc{ScreenHaystack}.}
Each heatmap shows grounding accuracy on the \(40\times40\) probing grid after aggregating across backgrounds and probe icons. Persistent low-accuracy regions indicate model-specific blind zones used for ScreenSpot-Pro validation and controlled relocation.}
   
    \label{fig:map}
\end{figure*}

\section{\textsc{ScreenHaystack} Benchmark}
\label{sec:screenhaystack}

Existing GUI grounding benchmarks evaluate static examples, where each target appears at a fixed location. This setting measures overall grounding performance, but does not reveal whether a model can localize the same target reliably across the image plane. \textsc{ScreenHaystack} exposes this missing spatial dimension by treating target location as a controlled variable. The benchmark is diagnostic rather than merely more difficult: its goal is not to introduce new visual or linguistic complexity, but to test whether the same grounding problem remains stable when relocated across the screen.

Specifically, each \textsc{ScreenHaystack} example is a visual needle-in-a-haystack task. We insert a synthetic target into a high-resolution GUI screenshot and ask the model to click the target. We use ten high-resolution UI-Vision~\citep{nayak2025ui} screenshots as backgrounds and three probe formats: pure icons, icons with text labels, and icons containing embedded text (Figure~\ref{fig:dataset}). The probing screenshots have a resolution of \(3360\times1924\) pixels. For each of the 30 background--format combinations, we probe a \(40\times40\) grid, yielding \(30\times40\times40=48{,}000\) probing instances in total. Each instance is assigned to a cell according to the target center. Since target identity and instruction are controlled, differences across grid cells mainly indicate location-dependent grounding reliability rather than variations in language, semantics, or interface content.

This construction separates \emph{what} must be grounded from \emph{where} it appears: the target description remains fixed while its position changes. 
It also converts grounding evaluation from a single score into a spatial reliability map, allowing persistent location-dependent failures to be distinguished from background- or icon-specific errors.

To evaluate a model on \textsc{ScreenHaystack}, we record the target-center grid cell for each probing instance and count a prediction as correct if the returned point falls inside the target region. We then pool results within each grid cell across all backgrounds and target formats to obtain a cell-wise accuracy map. 
Low-accuracy cells reveal spatial regions where the model is less reliable, which we call blind zones. We use two complementary blind-zone definitions. 
For downstream transfer and relocation experiments, we rank grid cells by probing accuracy and define the bottom \(20\%\) as the model-specific blind-zone mask, giving each model the same spatial budget. 
For diagnostic reporting, we instead use a fixed threshold \(B=\{i:a_i<0.8\}\), where \(a_i\) is the probing accuracy of cell \(i\). Blind-zone area \(R\) measures the fraction of cells in \(B\), blind-zone accuracy \(A_{\mathrm{blind}}\) measures the mean accuracy inside \(B\), relative accuracy drop \(F_{\mathrm{blind}}\) compares blind-zone accuracy with overall probing accuracy, and error concentration \(G_e\) measures how localized the errors are across the screen. Detailed formulas are provided in Appendix~\ref{app:blind_zone_metrics}.

\section{Models Exhibit Blind Zones}
\label{sec:blind-zone-existence}

This section asks whether current GUI grounding models exhibit spatially structured weaknesses. If grounding errors were mainly caused by semantic ambiguity, visual clutter, or instruction difficulty, then moving the same target across the screen should not produce consistent location-dependent failures. In contrast, if models have blind zones, their accuracy should drop systematically when targets appear in particular regions of the image plane. We test this hypothesis in three steps: first, we use \textsc{ScreenHaystack} to identify blind zones under the probing procedure; second, we examine whether the discovered blind zones also correspond to lower accuracy on a real ScreenSpot-Pro dataset; and third, we perform controlled relocation experiments to test whether moving the same target into or out of blind zones changes model accuracy.

\begin{table*}[t]
\centering
\footnotesize
\setlength{\tabcolsep}{13.5pt}
\renewcommand{\arraystretch}{1.12}
\begin{tabular}{lcccc}
\toprule
\textbf{Model} 
& \shortstack[c]{\textbf{Blind-zone area}\\\textbf{(\(a_i<0.8\))}} 
& \shortstack[c]{\textbf{Blind-zone}\\\textbf{accuracy}} 
& \shortstack[c]{\textbf{Relative}\\\textbf{accuracy drop}} 
& \shortstack[c]{\textbf{Error}\\\textbf{concentration}} \\
& \textbf{\(R\), \% cells $\downarrow$} 
& \textbf{\(A_{\mathrm{blind}}\), \% acc. $\uparrow$} 
& \textbf{\(F_{\mathrm{blind}}\), \% $\downarrow$} 
& \textbf{\(G_e\), Gini} \\
\midrule
Qwen3-VL-8B-Instruct  & \textbf{8.19}  & \textbf{70.53} & 23.67  & 0.538 \\
GTA1-7B               & 46.31 & 66.52 &  15.78 & 0.375 \\
Qwen3-VL-32B-Instruct & 45.63 & 19.46 & 68.27 & 0.583 \\
UI-TARS-1.5-7B        & 62.81 & 64.84 &  \textbf{11.74} & 0.306 \\
UI-Venus-Ground-7B    & 22.00 & 47.39 &  45.14 & 0.759 \\
\bottomrule
\end{tabular}
\caption{\textbf{Blind-zone severity on \textsc{ScreenHaystack}.}
Blind-zone cells are defined by a fixed probing-accuracy threshold, \(B=\{i:a_i<0.8\}\). 
\(R\) reports the fraction of such cells, \(A_{\mathrm{blind}}\) reports their mean probing accuracy, and \(F_{\mathrm{blind}}\) measures the relative accuracy drop from overall probing accuracy to blind-zone accuracy.
\(G_e\) measures whether cell-wise errors are spatially concentrated or diffuse; larger values indicate more localized failures.}
\label{tab:blind_zone_severity}
\end{table*}

\begin{figure*}[t]
    \centering
    \includegraphics[width=\linewidth]{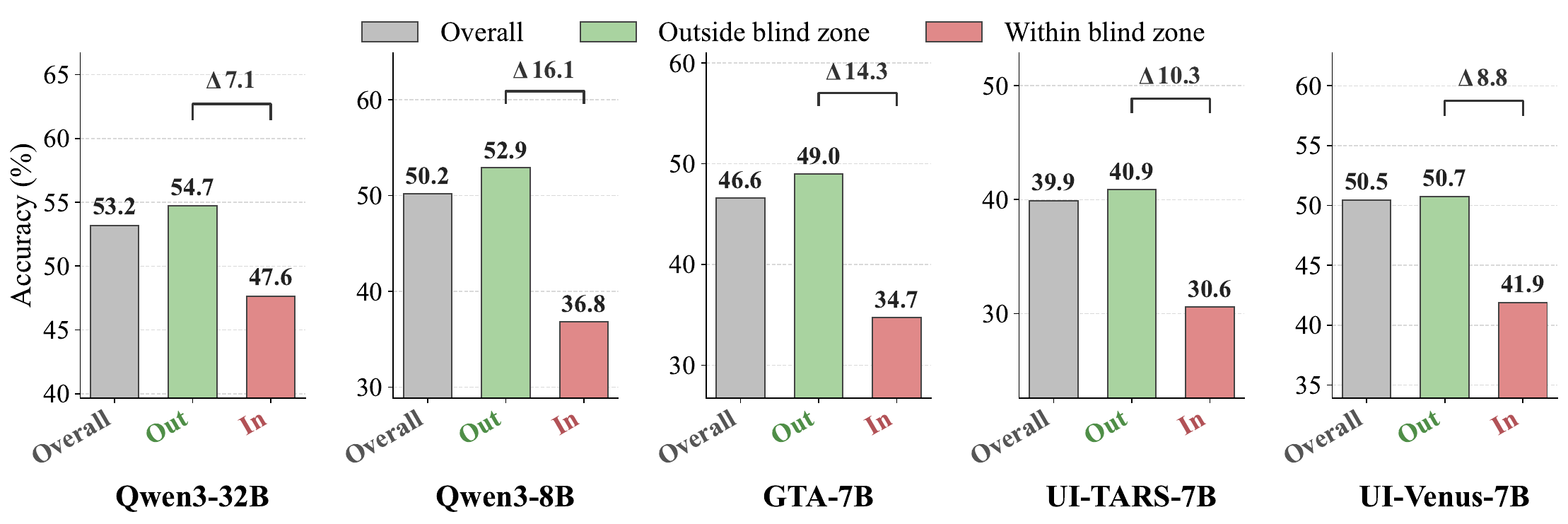}
    \caption{\textbf{Real-task validation of discovered blind zones.} We project the \textsc{ScreenHaystack}-derived blind-zone mask for each model onto unseen ScreenSpot-Pro samples and split targets by whether their centers fall inside or outside the mask. Across all evaluated models, targets inside blind zones have lower grounding accuracy than targets outside them. This shows that the probing-derived spatial maps predict failures on realistic GUI grounding examples, rather than only on synthetic probes.}
    \label{zone_results}
\end{figure*}
\textbf{GUI grounding models exhibit clear, model-specific blind zones.}
Figure~\ref{fig:map} shows that GUI grounding models exhibit blind zones. The spatial maps reveal structured low-accuracy regions for all evaluated models, including Qwen3-VL, GTA, UI-TARS, and UI-Venus, but their locations and shapes differ substantially. For example, Qwen3-VL-32B-Instruct has a pronounced weak band near the middle of the screen, while Qwen3-VL-8B-Instruct shows fewer and more scattered weak regions. GTA1-7B and UI-TARS-1.5-7B exhibit broader low-reliability regions toward the right side of the screen, whereas UI-Venus-Ground-7B shows a narrower vertical weak region. These patterns suggest that blind zones are structured and model-specific rather than uniformly distributed errors.

\textbf{Blind-zone diagnostics quantify extent and severity.}
Table~\ref{tab:blind_zone_severity} quantifies these differences with multiple diagnostics, since models differ in their overall grounding ability. Blind-zone area measures spatial extent: UI-TARS-1.5-7B has the broadest blind zones, with \(R=62.81\%\) of grid cells falling below the \(0.8\) accuracy threshold. Within the same Qwen3-VL family, blind-zone area does not appear to simply decrease with model scale: Qwen3-VL-8B-Instruct has the smallest area (\(R=8.19\%\)), while Qwen3-VL-32B-Instruct reaches \(R=45.63\%\). Qwen3-VL-32B-Instruct also has especially severe blind zones, with only \(19.46\%\) blind-zone accuracy and a \(68.27\%\) relative accuracy drop from overall probing accuracy. UI-Venus-Ground-7B has the most concentrated error pattern, with \(G_e=0.759\). These results show why a single overall accuracy number is incomplete: blind-zone maps reveal whether failures are broad, severe, concentrated, or comparatively mild.
    
\begin{figure*}[t]
    \centering
    \includegraphics[width=1\linewidth]{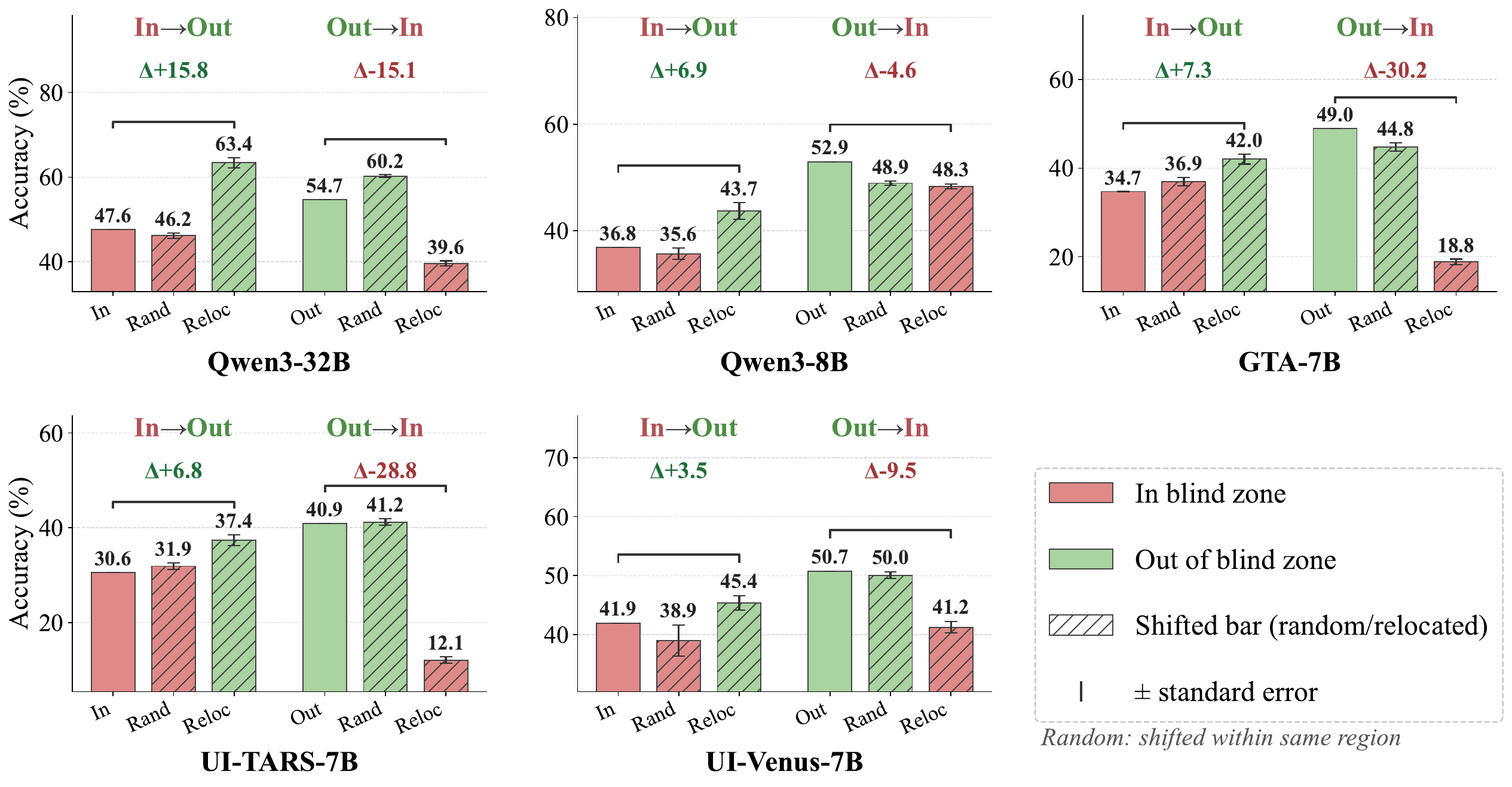}
    \caption{\textbf{Interventional validation via controlled spatial relocation.}
    Wrap-around shifts relocate the same ScreenSpot-Pro targets across the blind-zone boundary, while within-region random shifts serve as controls. For targets originally inside blind zones, relocating them out raises accuracy above the \emph{In} baseline; for targets originally outside, relocating them in lowers accuracy below the \emph{Out} baseline. Random-shift controls stay close to their baselines, showing the effect is driven by the blind-zone boundary rather than the shift itself. Error bars denote standard error; \(\Delta\) values are relocation effect sizes in percentage points.}
    \label{interventional_validation}
\end{figure*}

\textbf{Blind zones transfer to real-world GUI grounding data.}
We next test whether blind zones discovered by \textsc{ScreenHaystack} predict failures on ScreenSpot-Pro. 
As shown in Figure~\ref{zone_results}, targets whose centers fall inside a model's blind-zone mask are consistently harder to ground than targets outside the mask. 
Qwen3-VL-32B-Instruct drops from 54.7\% accuracy outside blind zones to 47.6\% inside blind zones, while Qwen3-VL-8B-Instruct drops from 52.9\% to 36.8\%. 
GTA1-7B decreases from 49.0\% to 34.7\%, UI-TARS-1.5-7B from 40.9\% to 30.6\%, and UI-Venus-Ground-7B from 50.7\% to 41.9\%. 
Across the five models, the average inside-versus-outside gap is 11.3 percentage points. 
This consistent degradation across model families indicates that the discovered blind zones capture a real spatial reliability pattern rather than probing noise. 
A simple location probe therefore identifies which realistic examples are harder to ground, indicating that future grounding evaluations should report spatially conditioned performance alongside overall accuracy. For Qwen3-VL-8B-Instruct, this gap also persists when ScreenSpot-Pro is stratified by element type, platform, and application domain (Appendix~\ref{app:stratified_screenspot}).

\textbf{Controlled relocation confirms a spatial effect.}
Finally, we test whether blind zones directly affect grounding accuracy by relocating ScreenSpot-Pro targets across the blind-zone boundary. As shown in Figure~\ref{interventional_validation}, we apply wrap-around shifts to the screenshot and target annotation so that the target moves to a new spatial region while preserving the original visual content and instruction. The detailed transformation procedure is provided in Appendix~\ref{app:wrap_around_shift}. Moving targets from outside to inside blind zones reduces accuracy for every model, with an average drop of 17.6 percentage points. The effect is largest for GTA1-7B, where accuracy drops by 30.2 points, and is similarly strong for UI-TARS-1.5-7B with a 28.8-point drop. Conversely, moving targets from inside to outside blind zones improves accuracy for every model, with an average gain of 8.1 points; the largest gain appears for Qwen3-VL-32B-Instruct, at 15.8 points. The within-region random-shift controls are much smaller: outside-to-outside shifts change accuracy by only -0.6 points on average, and inside-to-inside shifts by only -0.4 points. This contrast suggests that the effect is driven by crossing the blind-zone boundary rather than by image shifting itself. Blind zones therefore do not fully determine accuracy, but they are a measurable, transferable, and manipulable risk factor for GUI grounding.

\begin{findingbox}
\noindent\textbf{\textsc{Finding.}}
\textsc{ScreenHaystack} reveals model-specific blind zones that transfer to unseen ScreenSpot-Pro examples. Targets inside these regions are harder to ground, and controlled relocation shows that moving targets into blind zones lowers accuracy while moving them out improves accuracy.
\end{findingbox}

\begin{figure*}[ht]
    \centering
    \includegraphics[width=1\linewidth]{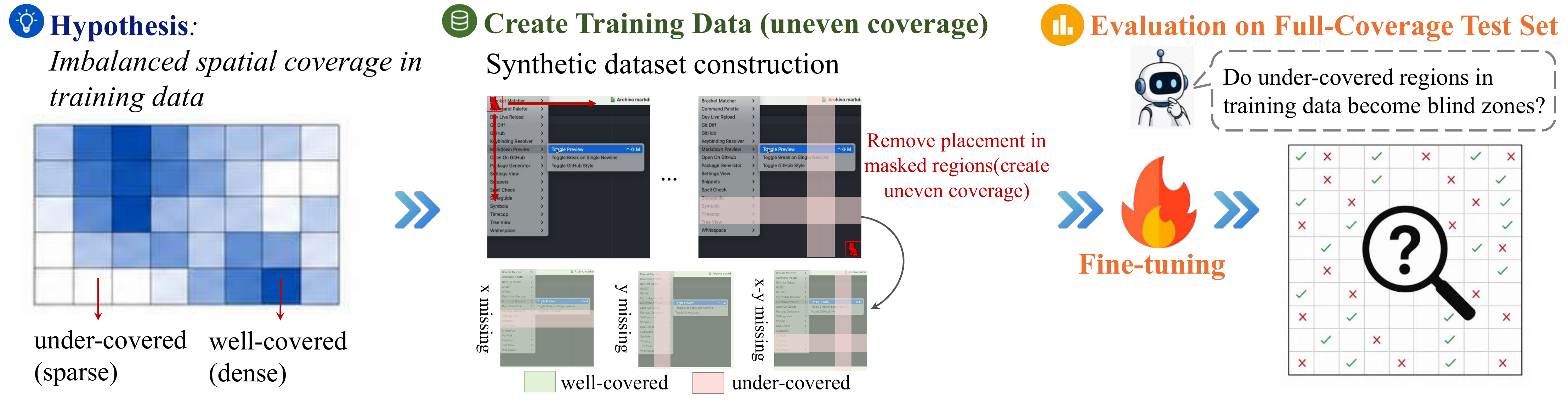}
    \caption{\textbf{Controlled synthetic experiment for testing whether uneven spatial coverage induces blind zones.} We create synthetic GUI grounding data with intentionally imbalanced target-center coverage by omitting training examples from specific spatial regions. The models are fine-tuned on these under-represented training sets and evaluated on a comprehensive test set. This approach enables us to determine whether areas lacking supervision during training become blind spots during evaluation.}
    
    \label{fig:synthetic}
\end{figure*}

\section{From Causes to Mitigation}

\subsection{What Causes Blind Zones?}
\label{sec:blind_cause}

Having established that blind zones exist and have measurable effects on real GUI grounding data, we next study one plausible cause: uneven spatial coverage in training data. If some regions of the image plane receive less grounding supervision during training, a model may learn weaker associations between visual evidence and output coordinates in those regions. This explanation is especially plausible for GUI data, where targets are not uniformly distributed: menus, toolbars, navigation panels, and dialog buttons occupy some regions far more often than others. Because most state-of-the-art GUI grounding models are trained on private or mixed data, we cannot directly inspect their full spatial training distributions. We therefore use a controlled synthetic setting to test whether uneven spatial coverage can be one cause of blind-zone-like failures.

\begin{table}[t]
    \centering
    \small
    \setlength{\tabcolsep}{3pt}
    \renewcommand{\arraystretch}{1.13}
    \begin{tabular}{llrrr}
    \toprule
    \textbf{Model} &
    \textbf{Training} &
    \textbf{$x$-miss.} &
    \textbf{$y$-miss.} &
    \textbf{$x$--$y$-miss.} \\
    \midrule
    \multirow{3}{*}{\begin{tabular}[c]{@{}l@{}}Phi-3.5-Vision\\Instruct\end{tabular}}
    & Original  & 76.86 & 79.30 & 53.90 \\
    & Augmented & 91.30 & 89.50 & 82.20 \\
    & Gain      & \textbf{+14.44} & \textbf{+10.20} & \textbf{+28.30} \\
    \cmidrule(lr){1-5}
    \multirow{3}{*}{\begin{tabular}[c]{@{}l@{}}Qwen3-VL-8B\\Instruct\end{tabular}}
    & Original  & 94.40 & 95.30 & 77.40 \\
    & Augmented & 97.30 & 97.90 & 94.10 \\
    & Gain      & \textbf{+2.90} & \textbf{+2.60} & \textbf{+16.70} \\
    \bottomrule
    \end{tabular}
    \caption{\textbf{Blind-zone-oriented augmentation in controlled synthetic settings.}
    We fine-tune each model on synthetic training data with supervision removed from selected \(x\), \(y\), or joint \(x\)--\(y\) regions, and evaluate on a full-coverage test set. Accuracy (\%) is reported for the original under-covered training data and for the augmented data, with gains measured as percentage-point improvements over the corresponding original setting.}
\label{tab:synthetic_augmentation_results}
\end{table}

\textbf{Experiment settings.}
As shown in Figure~\ref{fig:synthetic}, we construct a synthetic localization dataset where target appearance, instruction format, and background set are fixed, while the spatial distribution of training targets is explicitly controlled. Each example consists of a GUI background image with the same inserted icon and an instruction asking the model to click the icon. To simulate spatial under-coverage, we remove training examples whose target coordinates fall into predefined regions, while keeping these regions present during evaluation. We consider three missing-coverage settings: removing examples by the target's \(x\) coordinate, removing examples by the target's \(y\) coordinate, and removing examples by both \(x\) and \(y\), which creates a localized two-dimensional coverage gap. This comparison tests whether models can generalize from one coordinate dimension alone or require joint two-dimensional coverage. We fine-tune Qwen3-VL-8B-Instruct and Phi-3.5-Vision-Instruct~\citep{abdin2024phi} under each setting and evaluate them on a test set with full spatial coverage. 

\textbf{Results.}
Table~\ref{tab:synthetic_augmentation_results} shows that both models perform worst when training data are missing jointly along both coordinate dimensions. For Phi-3.5-Vision-Instruct, accuracy drops to 53.90\% in the joint \(x\)--\(y\) missing setting, compared with 76.86\% and 79.30\% under the \(x\)-only and \(y\)-only missing settings. Qwen3-VL-8B-Instruct shows the same pattern, dropping to 77.40\% under joint missing coverage, compared with 94.40\% and 95.30\% under the single-coordinate missing settings. The gap is much larger for joint missing coverage, suggesting that models can interpolate from supervision along one coordinate but struggle when a local two-dimensional region receives little direct supervision. This pattern is also stronger for the smaller Phi model, while Qwen3-VL-8B remains more robust but still shows a 16.70-point drop in the joint missing setting. These results indicate that spatial coverage is not only a dataset statistic but a learnable source of grounding reliability: GUI grounding data should be balanced not only over element semantics and interface domains, but also over target-center density across the screen plane.

\begin{findingbox}
\noindent\textbf{\textsc{Finding.}}
Controlled synthetic experiments show that uneven spatial coverage in training data can induce blind-zone-like failures. When a region lacks supervision along both \(x\) and \(y\) coordinates, grounding accuracy drops most sharply in that region.
\end{findingbox}

\subsection{How to Mitigate Blind Zones?}
\label{sec:mitigation}

If blind zones arise partly from uneven spatial coverage in training data, then they may be mitigated by adding supervision in those weak regions. We therefore propose \emph{blind-zone-oriented augmentation}, a targeted augmentation method that transforms training examples so that target centers move toward the model's identified blind zones. The transformation preserves the original instruction and target content, but changes the target location to rebalance spatial supervision.

\begin{table}[t]
    \centering
    \small
    \setlength{\tabcolsep}{11.5pt}
    \renewcommand{\arraystretch}{1.13}
    \begin{tabular}{lcr}
    \toprule
    \textbf{Setting} &
    \textbf{FT} &
    \textbf{Acc.} \\
    \midrule
    Vanilla Qwen3-VL-8B
    & -- & 50.28 \\
    Original Click-100k
    & \checkmark & 59.45 \\
    Random augmentation
    & \checkmark & 60.38 ± 0.41  \\  
    Blind-zone augmentation
    & \checkmark & \textbf{61.47 ± 0.29}  \\  
    \bottomrule
    \end{tabular}
    \caption{\textbf{Blind-zone-oriented augmentation on real-world Click-100k fine-tuning.}
    ScreenSpot-Pro accuracy is reported for the vanilla Qwen3-VL-8B model and for Qwen3-VL-8B models fine-tuned on Click-100k with different augmentation strategies.}
\label{tab:click100k_augmentation_results}
\end{table}

\begin{figure}[t]
    \centering

    \includegraphics[width=0.98\linewidth]{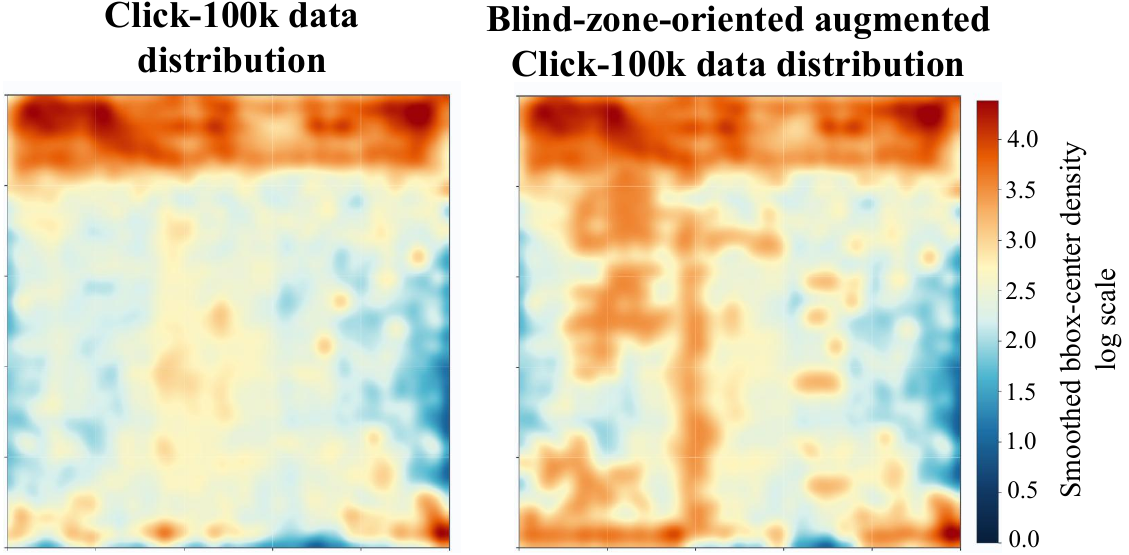}
   \caption{
\textbf{Click-100k data distribution before and after augmentation.}
The heatmaps show target-center density in normalized screen coordinates. Blind-zone-oriented augmentation increases coverage in sparse regions, yielding a more balanced training distribution.
}
    \label{fig:data_map}

    \vspace{0.6em}

    \includegraphics[width=0.98\linewidth]{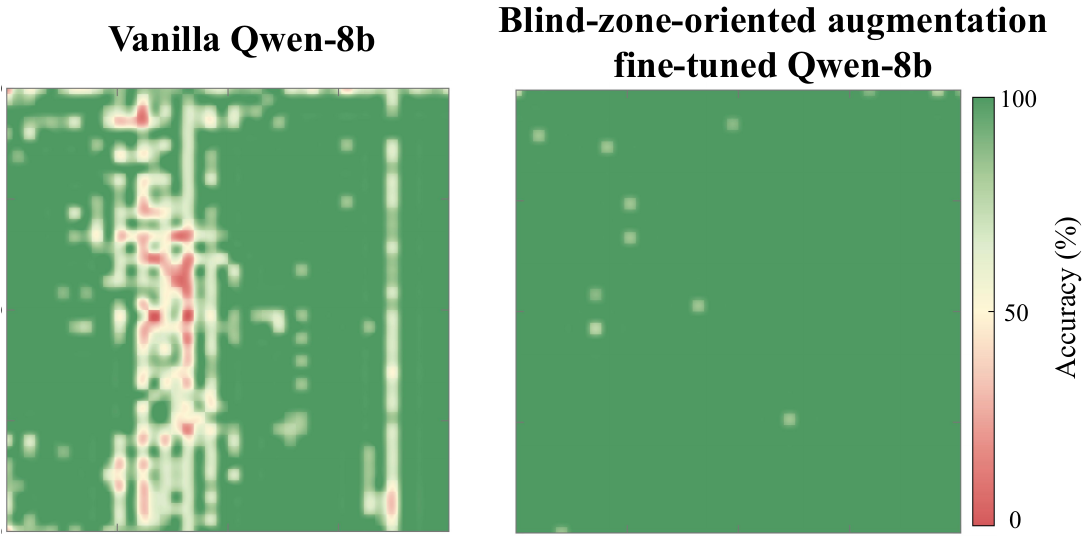}
 \caption{
\textbf{Spatial probing accuracy before and after blind-zone-oriented augmentation.}
On the same background and probe icon, blind-zone-oriented fine-tuning reduces low-accuracy regions.
}
    \label{fig:testzone}

\end{figure}
\textbf{Experimental settings.}
We evaluate blind-zone-oriented augmentation in both synthetic and real-world settings. In the synthetic setting, we augment each missing-coverage training set by adding transformed examples whose targets are moved into the under-covered regions. In the real-world setting, we fine-tune Qwen3-VL-8B-Instruct on Click-100k~\citep{gelato2025}, a GUI grounding dataset with approximately 101k examples. Figure~\ref{fig:data_map} shows that blind-zone-oriented augmentation increases target coverage in sparse spatial regions and yields a more balanced training distribution. We compare three training conditions: original Click-100k fine-tuning, random augmentation, and blind-zone-oriented augmentation. Both augmentation methods use the same crop-and-padding transformation and add the same number of examples; only blind-zone-oriented augmentation explicitly moves target centers into the model's identified blind zones. This baseline isolates whether \emph{where} additional supervision is placed matters beyond adding more transformed data.

\textbf{Results.} Tables~\ref{tab:synthetic_augmentation_results} and~\ref{tab:click100k_augmentation_results} show that targeted augmentation improves performance in both settings. In the synthetic setting, Phi-3.5-Vision-Instruct improves from 70.02\% to 87.67\% mean accuracy across the three missing-coverage settings, while Qwen3-VL-8B-Instruct improves from 89.03\% to 96.43\%. The largest gains appear in the joint \(x\)--\(y\) missing setting, where spatial under-coverage is most severe. In the real-world Click-100k setting, vanilla Qwen3-VL-8B-Instruct achieves 50.28\% accuracy on ScreenSpot-Pro. Fine-tuning on original Click-100k improves accuracy to 59.45\%.
Across three seeds, random augmentation achieves
\(60.38\pm0.41\%\), while blind-zone-oriented augmentation
achieves \(61.47\pm0.29\%\), outperforming random augmentation
in all three runs. Figure~\ref{fig:testzone} further shows this effect qualitatively on one fixed background and one fixed probe icon: the vanilla model exhibits obvious low-accuracy regions, whereas the blind-zone-oriented fine-tuned model shows substantially fewer blind zones. This is not a complete fix, but it shows that blind-zone maps can serve as feedback signals for dataset repair: diagnose where the model fails, add supervision there, and re-evaluate whether failures become less concentrated.

\begin{findingbox}
\noindent\textbf{\textsc{Finding.}}
Blind-zone-oriented augmentation mitigates blind zones by adding supervision in weak spatial regions. It improves synthetic missing-coverage performance and achieves the best ScreenSpot-Pro accuracy among the Click-100k fine-tuning settings.
\end{findingbox}

\subsection{Discussion}

\textbf{Are blind zones artifacts of synthetic probing?}
\textsc{ScreenHaystack} is designed to isolate spatial reliability rather than add visual or semantic difficulty. It uses three probe formats and multiple GUI backgrounds, and pools results across them so that blind zones reflect failures that persist across target appearances and interface contexts. Additional background-wise results in Appendix~\ref{app:additional_probing} show that the patterns are not driven by a single icon or background. The same masks also predict lower accuracy on unseen ScreenSpot-Pro examples (Figures~\ref{zone_results} and~\ref{interventional_validation}) and affect performance under controlled relocation, suggesting that blind zones are not merely synthetic-probe artifacts.

\noindent\textbf{Can architectural factors explain blind zones?}
We test two concrete architectural hypotheses on Qwen3-VL.

\emph{Visual-token boundaries.}
Qwen3-VL divides the processed image into merged visual tokens.
If blind zones were associated with visual-token boundaries, targets near these boundaries would be expected to exhibit lower grounding accuracy.
We reconstruct Qwen3-VL's effective visual-token lattice from the spatial token-grid dimensions produced by its vision processor.
With a patch size of \(16\) and a spatial merge factor of \(2\), adjacent merged visual tokens correspond to a \(32\)-pixel stride in the processed image.
We project these token boundaries back to the original ScreenSpot-Pro coordinate system and compare grounding accuracy for targets within \(8\) pixels of a boundary against those farther away.
Qwen3-VL-32B obtains \(53.4\%\) accuracy near boundaries and \(52.5\%\) farther away, while Qwen3-VL-8B obtains \(49.0\%\) and \(54.1\%\), respectively. These results suggest that boundary effects may exist for some models, but do not consistently account for the observed blind-zone patterns.

\emph{Coordinate-decoding drift.}
We also test whether blind zones arise from a systematic coordinate-decoding offset.
If this hypothesis were correct, prediction errors inside blind zones should point in a similar direction.
For each valid prediction, we compute the normalized error vector from the ground-truth target center to the predicted point and measure directional consistency as
\(\|\operatorname{mean}(\mathbf{e})\|_2 /
\operatorname{mean}(\|\mathbf{e}\|_2)\), where a value near \(1\) indicates a shared error direction and a value near \(0\) indicates dispersed directions.
For Qwen3-VL-8B-Instruct, the score is \(0.128\) inside blind zones, lower than \(0.247\) outside.
This result suggests that errors inside blind zones are more dispersed
rather than exhibiting a common directional shift.

Overall, these diagnostics suggest that visual-token boundaries and global coordinate drift alone may not fully explain blind zones.
Other factors, including preprocessing, architecture, training-data coverage, and alignment, may jointly contribute.

\section{Related Work}
    
\textbf{Computer-using agents and GUI grounding.}
GUI grounding is a core capability for computer-using agents: given a screenshot and an instruction, the model must localize the target UI element. Recent models and agents, including CogAgent~\citep{hong2024cogagent}, Ferret-UI~\citep{you2024ferret}, SeeClick~\citep{cheng2024seeclick}, UGround~\citep{gou2025navigating}, OS-Atlas~\citep{wu2025atlas}, UI-TARS~\citep{qin2025ui}, Aguvis~\citep{xu2024aguvis}, GUI-Actor~\citep{wu2026gui}, and GUI-G2~\citep{tang2026gui}, have improved localization on complex screenshots. Our work reveals a new failure mode in this setting: even when the target is visually salient and semantically simple, grounding accuracy can drop sharply in specific spatial regions of the image plane. We call these regions blind zones, study why they emerge, and propose a data-centric strategy to mitigate them.
        
\noindent\textbf{GUI grounding evaluation benchmarks.}
ScreenSpot and ScreenSpot-Pro~\citep{li2025screenspot} evaluate GUI localization across mobile, desktop, and web interfaces. UI-Vision~\citep{nayak2025ui}, MMBench-GUI~\citep{wang2025mmbenchgui}, and GroundCUA~\citep{feizi2026grounding} expand evaluation to high-resolution interfaces, diverse element types, and realistic GUI tasks. These benchmarks measure general grounding performance, but each target appears at a fixed position. \textsc{ScreenHaystack} instead dynamically places controlled targets across GUI backgrounds and reports spatially resolved accuracy.
        
\noindent\textbf{Robustness in GUI grounding.}
GUI grounding models can be brittle under changes to visual context or input presentation. MVP shows that cropped views can change coordinate predictions and improves inference-time stability with multi-view prediction~\citep{zhang2026mvp}. Other work refines search regions~\citep{nguyen2024improved,luo2025visual} or studies failures under attacks and domain perturbations~\citep{zhao2025robustness,wang2026gui}. These works show sensitivity to visual conditions; \textsc{ScreenHaystack} asks whether such failures concentrate in particular screen regions.

\section{Conclusion}

In summary, \textsc{ScreenHaystack} shifts GUI grounding evaluation from static examples to dynamic, location-controlled probing. This reveals blind zones: spatial regions where current models become systematically less reliable. We show that these blind zones transfer to real ScreenSpot-Pro examples and causally affect performance under controlled relocation: moving targets into them lowers accuracy, while moving targets out improves it. We further show that uneven spatial coverage in training data can induce such failures, and that blind-zone-oriented augmentation can mitigate them. These results highlight spatial reliability as a key axis for evaluating and training robust computer-using agents.


\section*{Limitations}

Our study has several limitations. First, \textsc{ScreenHaystack} uses controlled synthetic probes, which are intentionally simpler than natural UI elements. This design isolates spatial reliability, but it does not replace evaluation on real GUI tasks; we address this partly through ScreenSpot-Pro transfer and relocation experiments. 
Second, our data-centric experiments show that uneven spatial coverage is sufficient to induce blind-zone-like failures, but they do not prove that spatial coverage is the only cause of blind zones in deployed models. Other factors, including image resizing, positional encoding, attention allocation, and coordinate decoding, may also contribute. Finally, due to computational constraints, we evaluate a limited set of representative models and datasets; broader studies across architectures, scales, and training recipes would further clarify the generality of blind zones.

\bibliography{custom}

\appendix

\clearpage

\section{Blind-Zone Diagnostic Metrics}

\label{app:blind_zone_metrics}

We compute blind-zone diagnostic metrics from the probing accuracy map. 
Let \(a_i \in [0,1]\) denote the probing accuracy of grid cell \(i\), and let \(N\) be the total number of grid cells. 
For diagnostic reporting, we define blind-zone cells using a fixed probing-accuracy threshold:
\[
B=\{i:a_i<0.8\},
\]
and denote the non-blind-zone cells as \(\bar{B}=\{i:a_i\geq 0.8\}\).

\paragraph{Blind-zone area \(R\).}
Blind-zone area measures how widely blind zones spread across the screen:
\[
R=\frac{|B|}{N}.
\]
A larger \(R\) indicates that low-reliability regions occupy a broader spatial extent.

\paragraph{Blind-zone accuracy \(A_{\mathrm{blind}}\).}
Blind-zone accuracy measures the model's average reliability inside the identified blind-zone cells:
\[
A_{\mathrm{blind}}
=
\frac{1}{|B|}\sum_{i\in B} a_i.
\]
A lower \(A_{\mathrm{blind}}\) indicates that the model fails more often once the target falls inside the blind-zone regions.

\paragraph{Relative accuracy drop \(F_{\mathrm{blind}}\).}
We first compute the overall probing accuracy and the blind-zone internal accuracy:
\[
A_{\mathrm{orig}}
=
\frac{1}{N}\sum_{i=1}^{N}a_i,
\qquad
A_{\mathrm{blind}}
=
\frac{1}{|B|}\sum_{i\in B}a_i.
\]
The relative accuracy drop measures how much the blind-zone accuracy deviates from the overall probing accuracy in relative terms:
\[
F_{\mathrm{blind}}
=
\left|
\frac{A_{\mathrm{blind}}-A_{\mathrm{orig}}}{A_{\mathrm{orig}}}
\right|.
\]
Equivalently, since blind-zone accuracy is typically lower than the overall probing accuracy, this can be written as
\[
F_{\mathrm{blind}}
=
\frac{A_{\mathrm{orig}}-A_{\mathrm{blind}}}{A_{\mathrm{orig}}}.
\]
A larger \(F_{\mathrm{blind}}\) indicates stronger spatial unevenness: blind-zone regions are much less reliable than the model's overall average, suggesting stronger blind-zone-induced fluctuation.
\paragraph{Error concentration \(G_e\).}
Error concentration measures whether low-reliability behavior is localized in a few screen regions or diffusely spread across the image plane. 
We define the cell-wise blind-zone intensity as
\[
z_i=1-a_i.
\]
Then \(G_e\) is computed as the Gini coefficient of \(\{z_i\}_{i=1}^{N}\):
\[
G_e
=
\frac{
\sum_{i=1}^{N}\sum_{j=1}^{N}|z_i-z_j|
}{
2N^2\bar{z}
},
\qquad
\bar{z}=\frac{1}{N}\sum_{i=1}^{N}z_i.
\]
If \(\bar{z}=0\), we set \(G_e=0\). 
Larger values indicate that blind-zone behavior is more spatially localized, while smaller values indicate that low-reliability regions are more diffusely spread across the screen.

\section{Prompt Templates}
\label{app:prompt_templates}

This appendix reports the exact prompt templates used in our probing and test-set evaluations. 
All prompts are designed to standardize the grounding task across models by specifying the image resolution, requiring the target center for elements with spatial extent, and constraining the expected output to a single coordinate pair. 
This design reduces variation due to prompt format and allows all predictions to be evaluated under the same coordinate-level correctness criterion.
\subsection{Visual Probing Prompt}
\label{app:visual_probing_prompt}

For \textsc{ScreenHaystack} probing, we use a fixed locator prompt for models that directly output click coordinates. 
The prompt specifies the coordinate system, the image resolution, and the required output format. 
For area-like targets, models are instructed to return the center point of the target.

\begin{promptbox}{Visual probing prompt}
You are an expert UI element locator. Given a GUI image and a user's element description, provide the coordinates of the specified element as a single \((x,y)\) point. The image resolution is height \(\{H\}\) and width \(\{W\}\). For elements with area, return the center point.

\medskip
Output the coordinate pair exactly:

\[
(x,y)
\]

\medskip
Element description: \(\{\texttt{prompt\_text}\}\)
\end{promptbox}

For UI-Venus, which follows a bounding-box output convention, we use its native outline-style prompt in the visual probing experiments:

\begin{promptbox}{UI-Venus visual probing prompt}
Outline the position corresponding to the instruction: \(\{\texttt{element\_description}\}\). The output should be only [x1,y1,x2,y2].
\end{promptbox}

Here, \(\{H\}\) and \(\{W\}\) denote the height and width of the input screenshot, \(\{\texttt{prompt\_text}\}\) is replaced by the description of the inserted target, and \(\{\texttt{element\_description}\}\) denotes the same target description under the UI-Venus prompt format. 
For point-output models, the parsed \((x,y)\) coordinate is used directly as the predicted target location. 
For UI-Venus, we parse the predicted bounding box \([x_1,y_1,x_2,y_2]\) and convert it to its center point,
\[
\left(\frac{x_1+x_2}{2}, \frac{y_1+y_2}{2}\right).
\]
A prediction is counted as correct if the resulting point lies inside the ground-truth bounding box of the inserted target. 
Thus, although UI-Venus produces a bounding box, all models are evaluated under the same point-in-box correctness criterion.

We use three target formats in order to reduce dependence on a single visual appearance or textual cue:

\begin{description}
    \item[Pure icon.] 
    ``A red five-pointed star shape.''

    \item[Icon with text label.] 
    ``the OK icon, a red circle with white text `OK'.''

    \item[Icon with embedded text.] 
    ``the clock icon, a white circle with black hour hand pointing to 3 o'clock and minute hand pointing to 12 o'clock, with text `clock' on the right.''
\end{description}

Because the target identity, instruction template, and output format are held fixed within each probing condition, differences in accuracy across grid cells primarily reflect location-dependent grounding reliability rather than changes in task wording or target semantics.

\subsection{Test-set Evaluation Prompt}
\label{app:evaluation_prompt}

For real-world test-set evaluation, we use a unified point-output locator prompt for models that directly predict click coordinates. 
The prompt defines the coordinate system and output convention but does not include the sample-specific target description. 
Instead, the target instruction is provided as a separate text segment after the image, matching the multimodal input structure used in our implementation.

\begin{promptbox}{Common locator prompt for test-set evaluation}
You are an expert UI element locator. Given a GUI image and a user's element description, provide the coordinates of the specified element as a single \((x,y)\) point. The image resolution is height \(\{\texttt{height}\}\) and width \(\{\texttt{width}\}\). For elements with area, return the center point.

\medskip
Output the coordinate pair exactly:

\[
(x,y)
\]
\end{promptbox}

For each point-output model, the input is constructed as a three-part multimodal message:

\begin{promptbox}{Multimodal input structure}
\texttt{[common locator prompt]}

\medskip
\texttt{[GUI image]}

\medskip
\texttt{[sample-specific grounding instruction]}
\end{promptbox}

For UI-Venus, we use its native bounding-box evaluation prompt:

\begin{promptbox}{UI-Venus test-set evaluation prompt}
Outline the position corresponding to the instruction: \(\{\texttt{instruction}\}\). The output should be only [x1,y1,x2,y2].
\end{promptbox}

For UI-Venus, the sample-specific grounding instruction is inserted into the bounding-box prompt above. 
This preserves the model's expected output format while keeping the target instruction semantically consistent with the other models.

We use these prompts for the original ScreenSpot-Pro evaluation, the wrap-around relocation experiments, and the synthetic evaluation settings. 
Model responses are parsed into image coordinates before scoring. 
For point-output models, a prediction is counted as correct only if the parsed point lies inside the ground-truth target region. 
For UI-Venus, we parse the predicted bounding box \([x_1,y_1,x_2,y_2]\), convert it to its center point,
\[
\left(\frac{x_1+x_2}{2}, \frac{y_1+y_2}{2}\right),
\]
and apply the same point-in-box correctness criterion. 
Responses that cannot be parsed into a valid coordinate pair or bounding box are counted as incorrect, including responses with missing coordinates, multiple ambiguous predictions, or natural-language outputs that do not follow the required format.

\section{Wrap-around Shift Transformation}
\label{app:wrap_around_shift}

For controlled spatial relocation, we use a wrap-around shift transformation.
Given an input screenshot $I$ with width $W$ and height $H$, we shift all pixels by an offset $(\Delta x, \Delta y)$.
Pixels that move beyond one image boundary reappear from the opposite boundary.
Formally, a pixel originally at $(x,y)$ is moved to
\[
\begin{aligned}
x' &= (x + \Delta x) \bmod W, \\
y' &= (y + \Delta y) \bmod H .
\end{aligned}
\]

The target bounding box is shifted by the same offset.
If the original target center is $(x_c,y_c)$, the shifted target center is
\[
\begin{aligned}
x_c' &= (x_c + \Delta x) \bmod W, \\
y_c' &= (y_c + \Delta y) \bmod H .
\end{aligned}
\]

This transformation preserves the full image content while changing the absolute image-plane location of the target.
It therefore allows us to compare model performance on the same grounding task before and after relocation.

\begin{figure*}[t]
    \centering
    \includegraphics[width=0.95\linewidth]{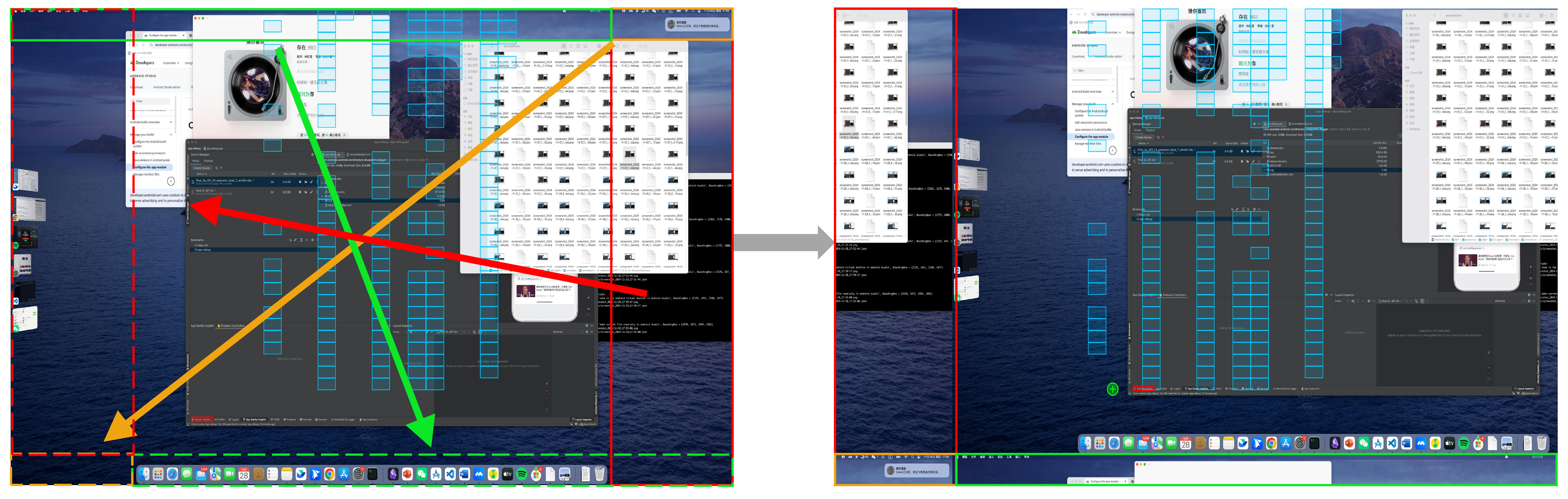}
    \caption{Illustration of the wrap-around shift transformation. The full screenshot is translated by a fixed offset, and pixels crossing one boundary reappear from the opposite boundary. The target bounding box is shifted by the same offset, so the target moves to a new spatial location while the image content is preserved.}
    \label{fig:app_wrap_around}
\end{figure*}

\begin{figure*}
    \centering
    \includegraphics[width=1\linewidth]{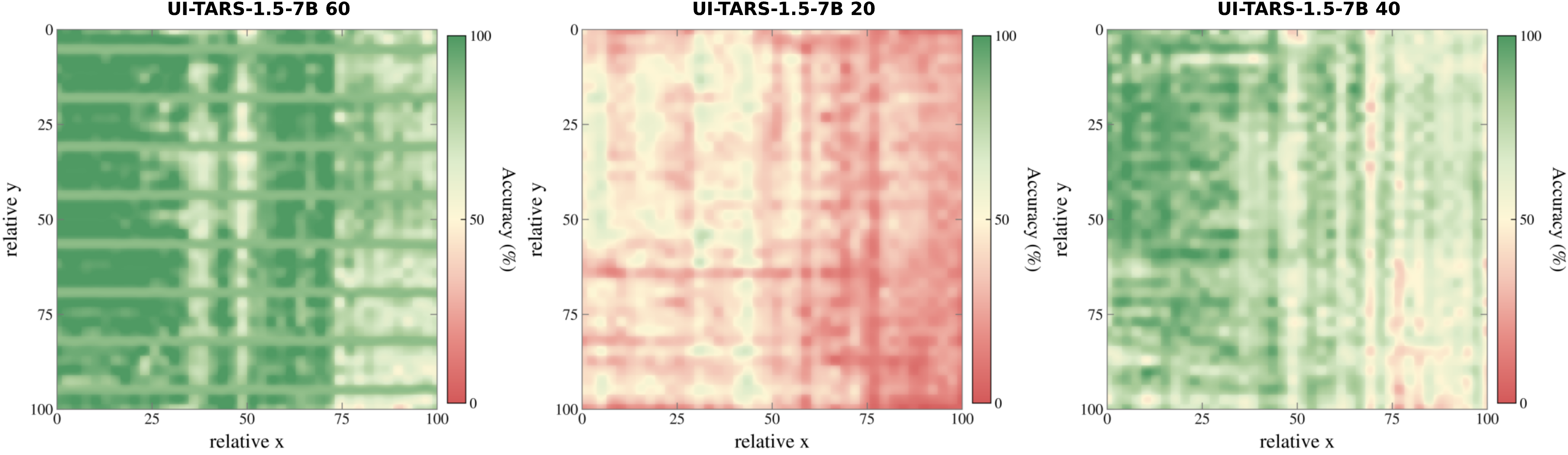}
    \caption{
\textbf{Representative probe icon size calibration for UI-TARS-1.5-7B.}
We show example spatial probing accuracy maps from the icon-size calibration sweep. 
Small targets lead to broad low-accuracy regions, suggesting that failures are partly driven by target visibility and recognition difficulty. 
Overly large targets make the task close to saturated, leaving weak spatial contrast. 
The intermediate scale preserves clear location-dependent variation while keeping the target recognizable, and is therefore used for the main \textsc{ScreenHaystack} probing experiments.
}
    \label{fig:size_calibration_uitars}
\end{figure*}
\section{Probe Icon Size Calibration}
\label{app:size_calibration}

We conduct an additional size-calibration experiment to ensure that the probing task measures location-dependent grounding reliability rather than trivial visibility or scale effects. The target icon must be large enough to be recognized by the model, but not so large that the localization task becomes saturated.

This calibration is important because very small targets can introduce a different failure mode from blind zones. When the model cannot reliably perceive the inserted icon, especially under greedy decoding, its output may collapse to a deterministic default behavior: it may repeatedly predict the same coordinate across many positions, or directly state that ``There is no \{icon\} visible in the provided image.'' Such failures reflect target-recognition difficulty rather than spatial grounding reliability, and would therefore confound the blind-zone analysis. Conversely, overly large icons make the task too easy because the target is visually salient and covers a large clickable area, causing most positions to be predicted correctly and weakening the diagnostic contrast across screen locations.

To choose an appropriate target scale, we conduct a size-sweep calibration using UI-TARS-1.5-7B. 
For the pure-icon and icon-with-text-label formats, we evaluate target sizes starting from small \(20\times20\) probes and increasing to larger candidates, including the \(40\times40\) scale used in the main benchmark. 
For the embedded-text format, which has a wider target region due to the text contained inside the icon, we evaluate a corresponding range from \(20\times30\) to \(60\times90\) pixels. 
Across this sweep, we select an intermediate size that keeps the target recognizable while avoiding saturation of the grounding task.

Figure~\ref{fig:size_calibration_uitars} shows representative examples from this calibration sweep for UI-TARS-1.5-7B. 
Small targets produce widespread failures across the image plane, indicating that the probing task can become dominated by target visibility and recognition difficulty rather than spatial grounding reliability. 
In contrast, overly large targets make the task substantially easier, with most regions reaching high accuracy and only weak spatial contrast remaining. 
The intermediate scale retains meaningful location-dependent variation without making the target difficult to recognize. 
This supports our choice of using an intermediate icon scale for the main probing benchmark.

For most evaluated models, we use \(40\times40\) pixels for the pure-icon and icon-with-text-label formats, and \(40\times60\) pixels for the embedded-text format. 
For Qwen3-VL-8B-Instruct, we use \(24\times24\) pixels for the pure-icon and icon-with-text-label formats, and \(45\times30\) pixels for the embedded-text format. 
These choices follow the same principle: the probe should be sufficiently visible for the model to identify the target, while remaining small enough to reveal meaningful spatial variation in grounding accuracy.

\section{Click-100k Augmentation Statistics}
\label{app:click100k_stats}

For real-data fine-tuning, we use Click-100k as the source training dataset. 
The original dataset contains approximately 101k GUI grounding examples, each consisting of a computer-screen image, a low-level grounding instruction, and a target annotation. 
We convert each target bounding box into its center point and normalize the resulting coordinate to the Qwen3-VL \texttt{norm1000} format.

We compare three fine-tuning settings. 
The first setting uses the original Click-100k data without additional transformations. 
The second setting adds randomly transformed examples as an augmentation baseline. 
The third setting applies blind-zone-oriented augmentation, where transformed examples are constructed so that their target centers are moved toward the previously identified blind zones of Qwen3-VL-8B-Instruct. 
The random and blind-zone-oriented augmentation settings add the same number of transformed examples, allowing us to isolate the effect of where the additional supervision is placed rather than the effect of increasing the training-set size.

Table~\ref{tab:app_click100k_stats} reports the corresponding dataset sizes. 
Both augmentation settings add 23,392 transformed examples, resulting in approximately 124.4k total examples for each augmented training set.

\begin{table*}[t]
\centering
\begin{tabular}{lccc}
\toprule
\textbf{Training setting} & \textbf{Original examples} & \textbf{Augmented examples} & \textbf{Total examples} \\
\midrule
Original Click-100k & $\sim$101k & 0 & $\sim$101k \\
Random augmentation & $\sim$101k & 23,392 & $\sim$124.4k \\
Blind-zone-oriented augmentation & $\sim$101k & 23,392 & $\sim$124.4k \\
\bottomrule
\end{tabular}
\caption{
Dataset sizes for Click-100k fine-tuning. 
Both augmentation settings add the same number of transformed examples, allowing random augmentation and blind-zone-oriented augmentation to be compared under the same data scale.
}
\label{tab:app_click100k_stats}
\end{table*}

\section{Fine-tuning Details}
\label{app:finetuning_details}

We fine-tune Qwen3-VL-8B-Instruct with LLaMA-Factory using LoRA-based supervised fine-tuning. 
For Click-100k, we convert each ground-truth bounding box into a target click point by taking the bounding-box center and mapping it to the Qwen3-VL \texttt{norm1000} coordinate system. 
The same preprocessing and optimization configuration is used for all Click-100k fine-tuning settings, including original fine-tuning, random augmentation, and blind-zone-oriented augmentation, so that performance differences can be attributed to the training data rather than to changes in the optimization setup.

All experiments are conducted in the supervised fine-tuning stage with LoRA adapters. 
We set the LoRA rank to \(8\) and apply LoRA to all target modules. 
The maximum sequence length is set to \(2048\), and the maximum number of image pixels is set to \(16{,}777{,}216\), allowing high-resolution GUI screenshots to be processed without aggressive downsampling. 
We use a per-device batch size of \(4\) with \(8\) gradient accumulation steps. 
The model is trained for \(2\) epochs with a learning rate of \(2.0\times10^{-4}\), a cosine learning-rate schedule, and a warmup ratio of \(0.1\). 
Training is performed in bfloat16 precision.

For data processing, we use a preprocessing batch size of \(32\), \(8\) preprocessing workers, and \(2\) dataloader workers. 
During training, we log metrics every \(10\) steps and save checkpoints every \(200\) steps. 
All reported fine-tuning results use this same configuration unless otherwise stated.

\section{Additional \textsc{ScreenHaystack} Probing Results}
\label{app:additional_probing}

We provide additional \textsc{ScreenHaystack} probing results to examine whether the discovered spatial reliability patterns are robust to variations in background interfaces and benchmark design choices.
A potential concern is that apparent blind zones may be induced by particular GUI layouts, local background clutter, or the visual appearance of a specific inserted target.
To reduce these confounding effects, our main blind-zone discovery procedure aggregates probing outcomes over multiple backgrounds and three probe formats, using repeated observations across diverse visual conditions to estimate each model's spatial reliability map.
This aggregation reduces the influence of isolated background- or probe-specific artifacts.

In this appendix, we further evaluate generalization across unseen backgrounds, sensitivity to grid resolution and blind-zone mask fraction, and stratified transfer on ScreenSpot-Pro.
We also report background-wise probing results for the original ten GUI screenshots.
Together, these analyses examine whether the observed blind-zone patterns depend on a particular background set, benchmark configuration, or ScreenSpot-Pro subgroup.
Across these analyses, the spatial reliability differences remain broadly consistent, supporting the interpretation that blind zones reflect model-specific spatial reliability patterns rather than artifacts of a particular probing configuration.

\subsection{Generalization Across Backgrounds}
\label{app:background_generalization}

We repeat the full probing procedure for UI-Venus-Ground-7B on ten additional unseen GUI backgrounds. The bottom-\(20\%\) blind-zone mask derived from these backgrounds has a Jaccard overlap of \(82.34\%\) with the original mask. As shown in Table~\ref{tab:background_generalization}, the two masks also predict nearly identical ScreenSpot-Pro gaps.

\begin{table}[t]
\centering
\small
\setlength{\tabcolsep}{4pt}
\begin{tabular}{lccc}
\toprule
\textbf{Mask source}
& \textbf{Outside}
& \textbf{Inside}
& \textbf{Gap} \\
\midrule
Original 10 backgrounds
& 50.7
& 41.9
& 8.8 \\
10 different backgrounds
& 51.5
& 42.9
& 8.6 \\
\bottomrule
\end{tabular}
\caption{\textbf{Generalization across backgrounds for UI-Venus-Ground-7B.}
We report ScreenSpot-Pro accuracy (\%) outside and inside each bottom-\(20\%\) blind-zone mask. Gap is Outside minus Inside in percentage points.}
\label{tab:background_generalization}
\end{table}

The high mask overlap and similar downstream gaps indicate that the spatial pattern is not specific to the original ten backgrounds.

\subsection{Sensitivity to Grid Resolution}
\label{app:grid_sensitivity}

We evaluate Qwen3-VL-8B-Instruct using \(40\times40\), \(50\times50\), and \(60\times60\) grids while keeping the bottom-\(20\%\) blind-zone mask definition fixed. Table~\ref{tab:grid_sensitivity} reports the resulting ScreenSpot-Pro gaps.

\begin{table}[t]
\centering
\scriptsize
\setlength{\tabcolsep}{3pt}
\begin{tabular}{lcccc}
\toprule
\textbf{Grid}
& \textbf{Cell size}
& \textbf{Outside}
& \textbf{Inside}
& \textbf{Gap} \\
\midrule
\(40\times40\)
& \(84.0\times48.1\)
& 52.9
& 36.8
& 16.1 \\
\(50\times50\)
& \(67.2\times38.5\)
& 52.4
& 40.2
& 12.2 \\
\(60\times60\)
& \(56.0\times32.1\)
& 52.1
& 39.7
& 12.4 \\
\bottomrule
\end{tabular}
\caption{\textbf{Sensitivity to grid resolution for Qwen3-VL-8B-Instruct.}
Cell sizes are in pixels. Outside and Inside are ScreenSpot-Pro accuracies (\%), and Gap is their difference in percentage points.}
\label{tab:grid_sensitivity}
\end{table}

The exact estimated boundaries vary with grid resolution, but the inside--outside gap remains substantial under all three settings.

\subsection{Sensitivity to Blind-Zone Mask Fraction}
\label{app:mask_sensitivity}

We vary the fraction of cells included in the operational blind-zone mask for Qwen3-VL-8B-Instruct while keeping the \(40\times40\) grid fixed. Table~\ref{tab:mask_sensitivity} reports the results.

\begin{table}[t]
\centering
\small
\setlength{\tabcolsep}{5pt}
\begin{tabular}{lccc}
\toprule
\textbf{Mask}
& \textbf{Outside}
& \textbf{Inside}
& \textbf{Gap} \\
\midrule
Bottom \(10\%\)
& 52.2
& 33.1
& 19.1 \\
Bottom \(20\%\)
& 52.9
& 36.8
& 16.1 \\
Bottom \(30\%\)
& 53.6
& 42.3
& 11.3 \\
\bottomrule
\end{tabular}
\caption{\textbf{Sensitivity to blind-zone mask fraction for Qwen3-VL-8B-Instruct.}
Outside and Inside are ScreenSpot-Pro accuracies (\%), and Gap is their difference in percentage points.}
\label{tab:mask_sensitivity}
\end{table}

The bottom-\(10\%\) mask isolates the most severe cells and produces the largest gap, while broader masks include more moderately reliable cells and reduce the contrast.

\begin{table}[t]
\centering
\scriptsize
\setlength{\tabcolsep}{3.5pt}
\renewcommand{\arraystretch}{1.05}
\begin{tabular}{lccc}
\toprule
\textbf{Stratum}
& \textbf{Outside}
& \textbf{Inside}
& \textbf{Gap} \\
\midrule
\textbf{Overall} & 52.90 & 36.80 & 16.10 \\
\midrule
\multicolumn{4}{l}{\textit{Element type}} \\
Icon & 22.98 & 13.04 & 9.94 \\
Text & 69.14 & 62.00 & 7.14 \\
\midrule
\multicolumn{4}{l}{\textit{Platform}} \\
macOS & 56.55 & 47.69 & 8.86 \\
Windows & 50.39 & 31.51 & 18.88 \\
\midrule
\multicolumn{4}{l}{\textit{Application}} \\
CAD & 44.70 & 40.54 & 4.16 \\
Creative & 46.33 & 37.14 & 9.19 \\
Development & 47.10 & 36.84 & 10.26 \\
Operating system & 52.05 & 19.05 & 33.00 \\
Numerical computing & 65.17 & 42.86 & 22.31 \\
Scientific plotting & 34.48 & 13.89 & 20.59 \\
Econometrics & 93.33 & 60.00 & 33.33 \\
Statistical analysis & 47.37 & 8.33 & 39.04 \\
\bottomrule
\end{tabular}

\caption{\textbf{Stratified ScreenSpot-Pro analysis for Qwen3-VL-8B-Instruct.}
Outside and Inside are accuracies (\%); Gap is Outside minus Inside in percentage points.}
\label{tab:stratified_screenspot}
\end{table}

\subsection{Stratified ScreenSpot-Pro Analysis}
\label{app:stratified_screenspot}

We stratify the Qwen3-VL-8B-Instruct ScreenSpot-Pro results by element type, platform, and application domain using the same \(40\times40\) grid and bottom-\(20\%\) blind-zone mask as in the main experiment. Table~\ref{tab:stratified_screenspot} reports the per-stratum accuracies.

Accuracy is lower inside blind zones for every reported element type, platform, and application group, indicating that the overall gap is not solely explained by one ScreenSpot-Pro subgroup.

\subsection{Background-wise Results with Probe Formats Aggregated}
\label{app:bg_aggregated_results}

\begin{figure*}
    \centering
    \includegraphics[width=1\linewidth]{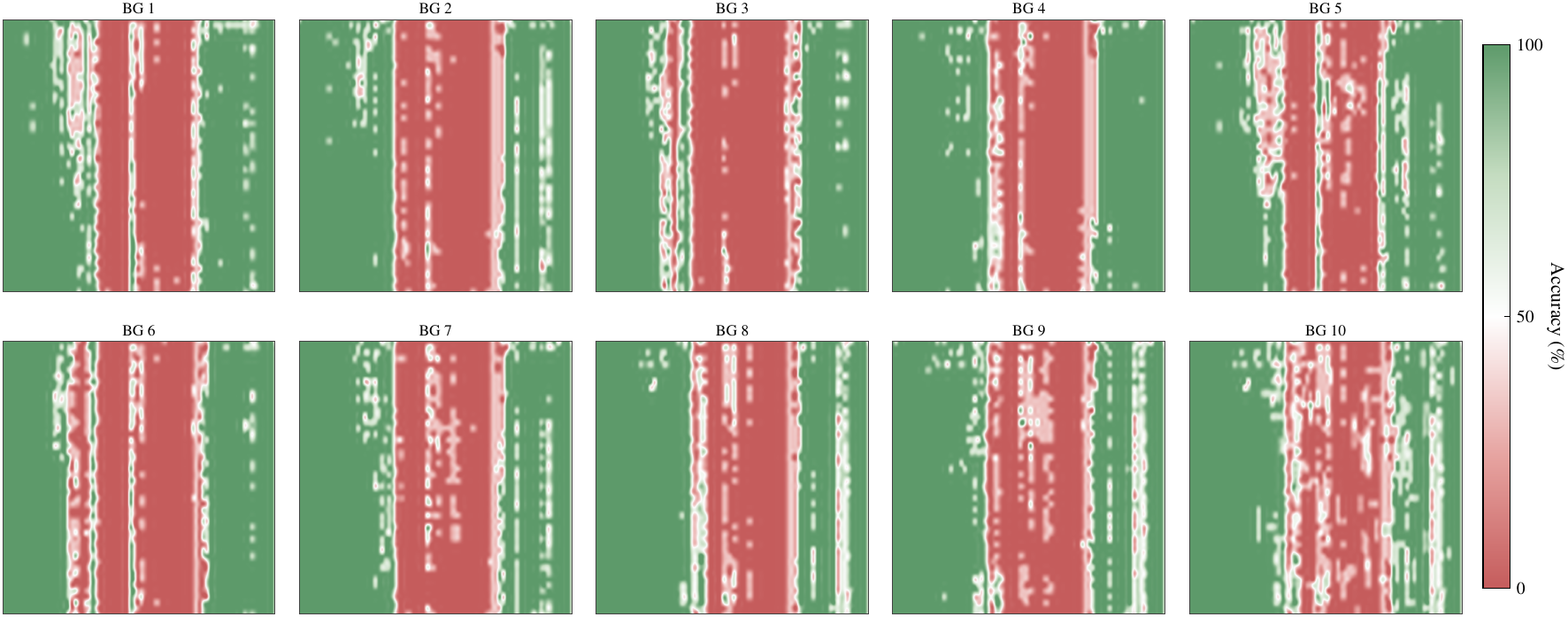}
    \caption{Background-wise \textsc{ScreenHaystack} accuracy maps for Qwen3-VL-32B-Instruct. For each background, accuracy is computed on the \(40\times40\) probing grid by aggregating the three probe formats.}
    \label{fig:app_qwen32b_bg_merged}
\end{figure*}

\begin{figure*}
    \centering
    \includegraphics[width=1\linewidth]{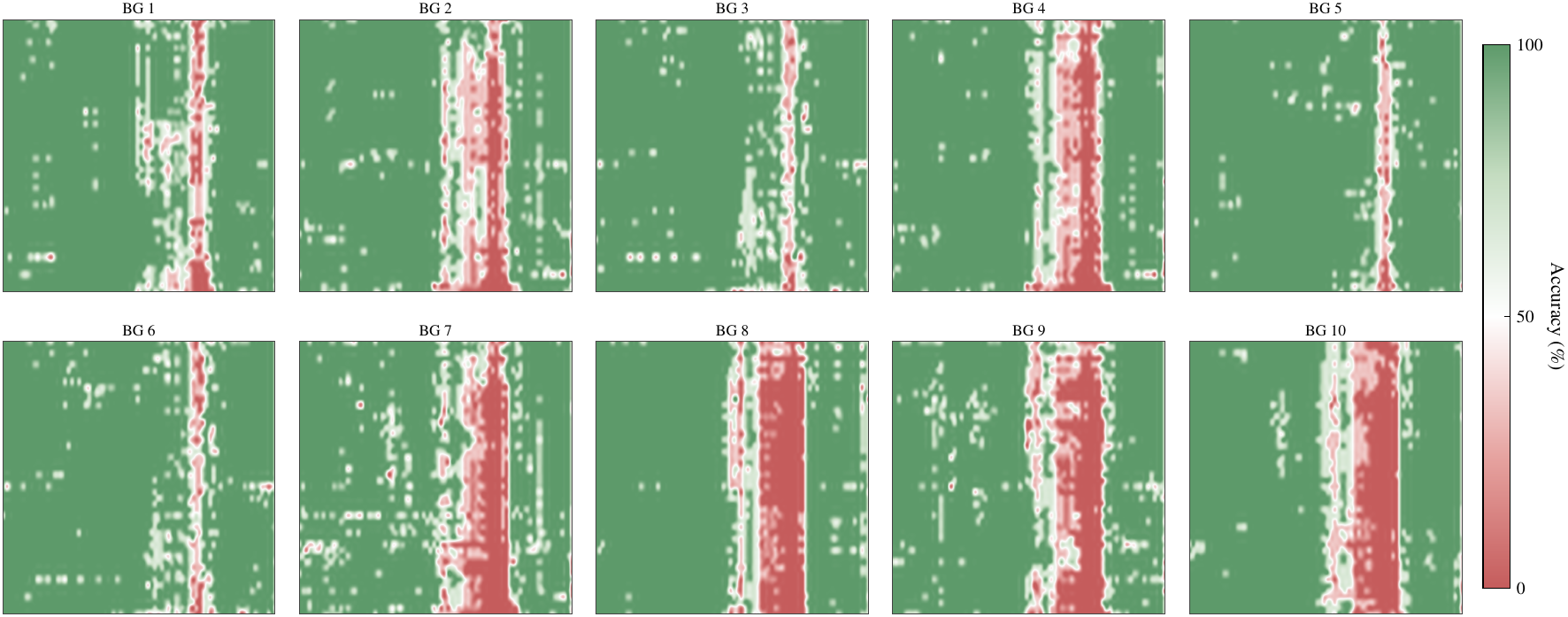}
    \caption{Background-wise \textsc{ScreenHaystack} accuracy maps for UI-Venus-Ground-7B. For each background, accuracy is computed on the \(40\times40\) probing grid by aggregating the three probe formats.}
    \label{fig:app_venus_bg_merged}
\end{figure*}

\begin{figure*}
    \centering
    \includegraphics[width=1\linewidth]{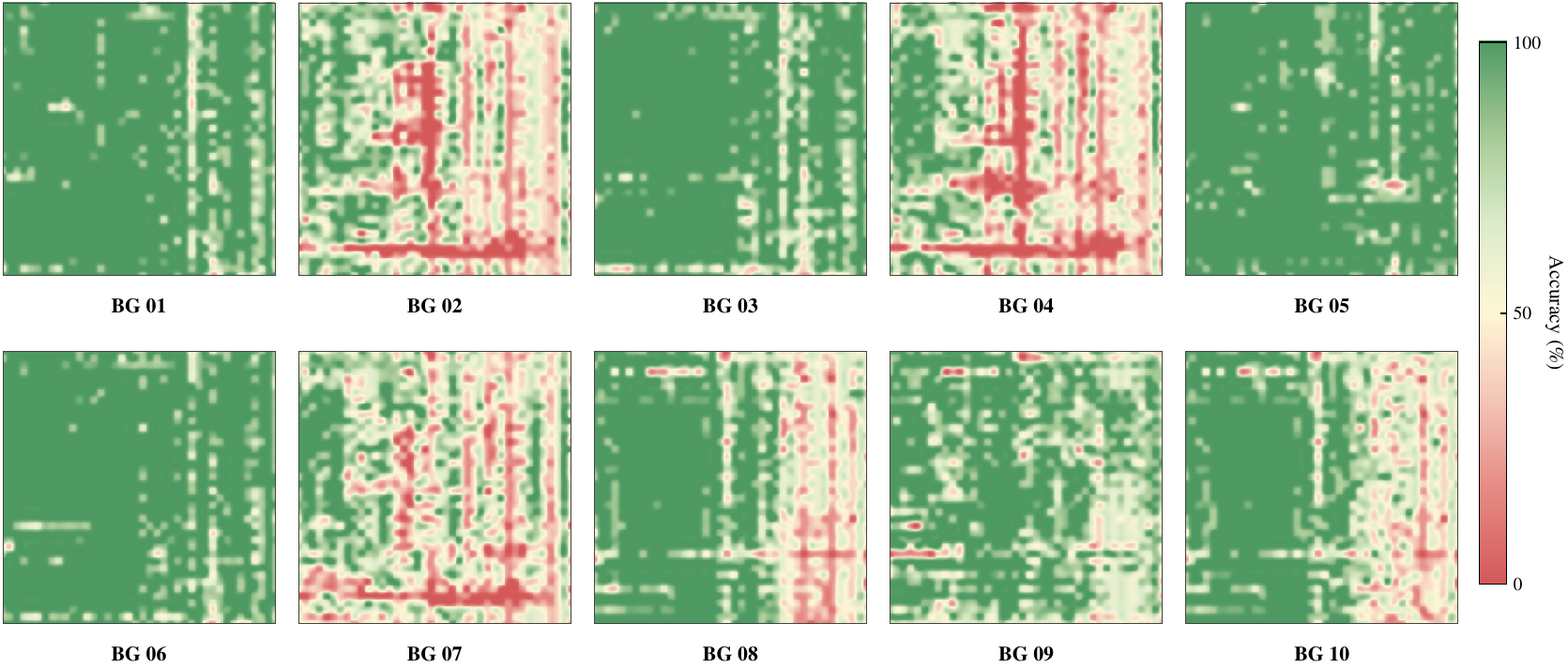}
    \caption{Background-wise \textsc{ScreenHaystack} accuracy maps for GTA1-7B. For each background, accuracy is computed on the \(40\times40\) probing grid by aggregating the three probe formats.}
    \label{fig:app_gta_bg_merged}
\end{figure*}

\begin{figure*}
    \centering
    \includegraphics[width=1\linewidth]{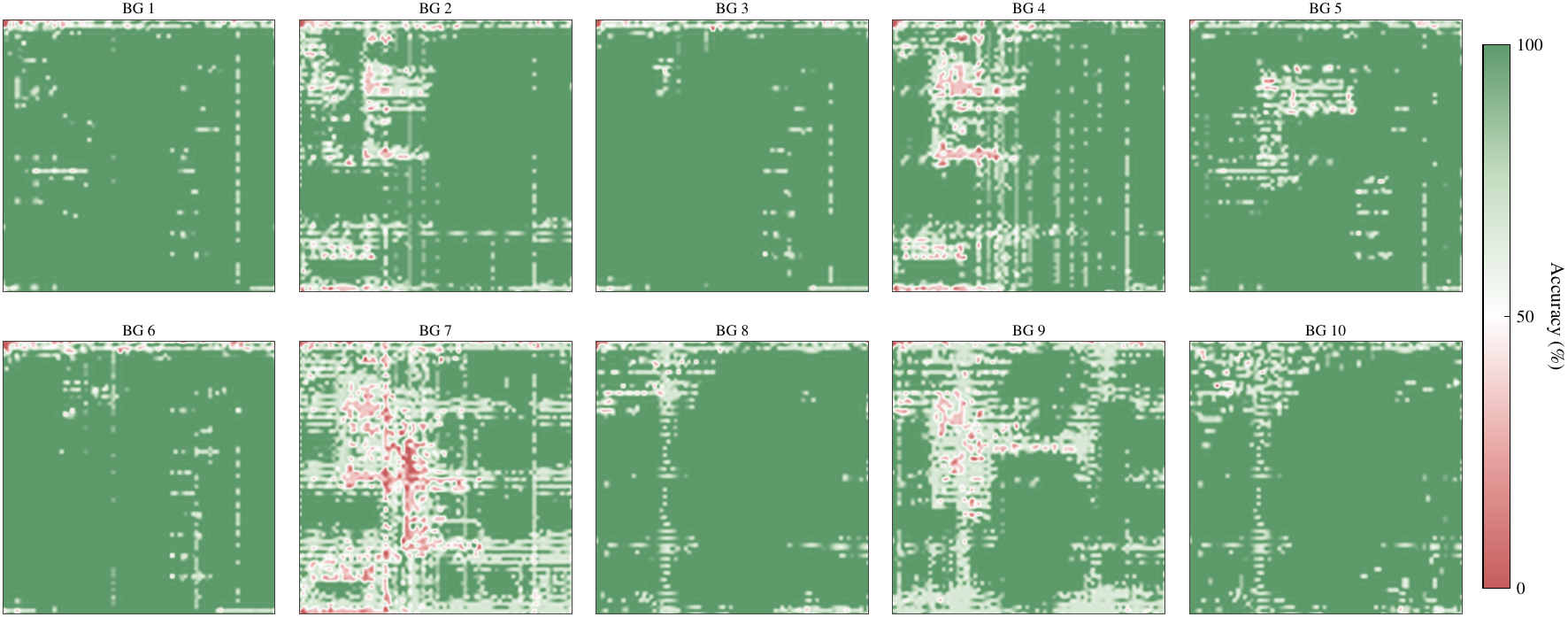}
    \caption{Background-wise \textsc{ScreenHaystack} accuracy maps for Qwen3-VL-8B-Instruct. For each background, accuracy is computed on the \(40\times40\) probing grid by aggregating the three probe formats.}
    \label{fig:app_qwen8b_bg_merged}
\end{figure*}

\begin{figure*}
    \centering
    \includegraphics[width=1\linewidth]{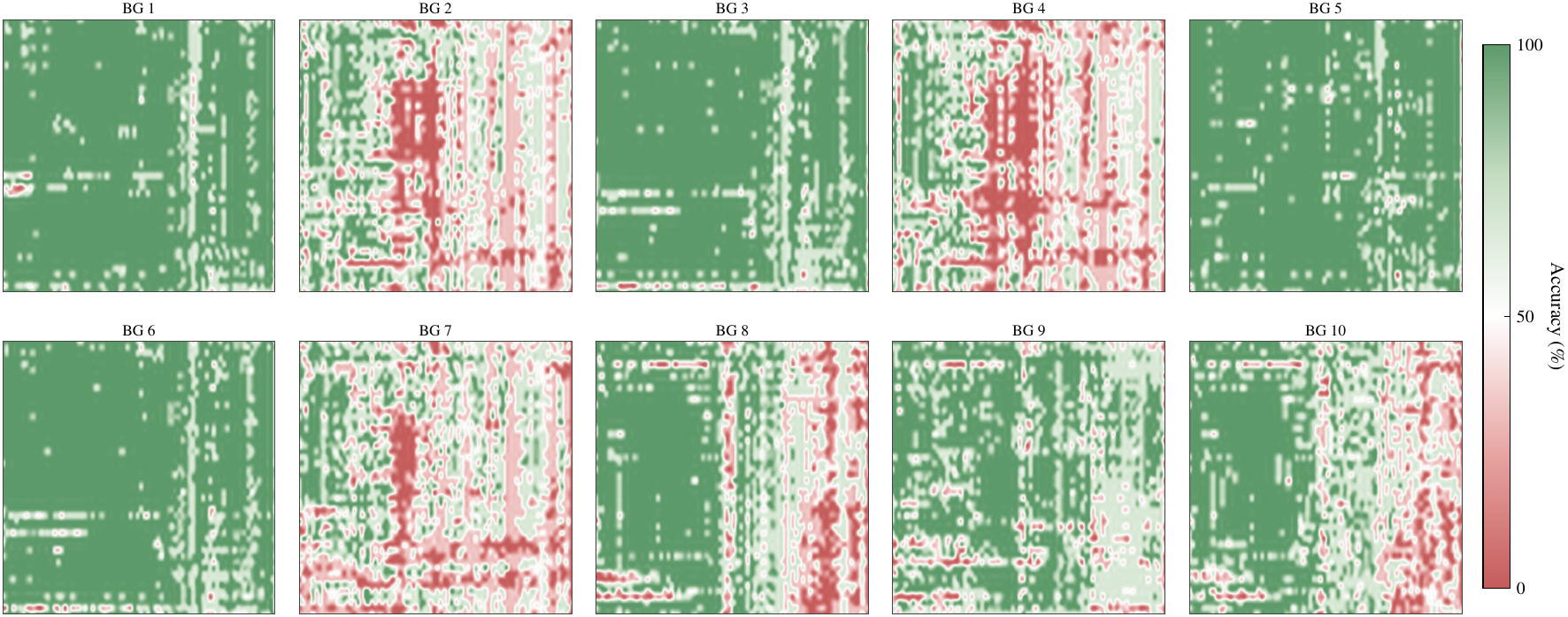}
    \caption{Background-wise \textsc{ScreenHaystack} accuracy maps for UI-TARS-1.5-7B. For each background, accuracy is computed on the \(40\times40\) probing grid by aggregating the three probe formats.}
    \label{fig:app_uitars_bg_merged}
\end{figure*}

Figures~\ref{fig:app_qwen32b_bg_merged}--\ref{fig:app_uitars_bg_merged} show background-wise \textsc{ScreenHaystack} accuracy maps for each evaluated model. 
For each background, accuracy is computed on the \(40\times40\) probing grid after pooling results from the three probe formats: Pure icon, Icon with text label, and Icon with embedded text. 
These figures provide a background-level view of each model's spatial reliability under different GUI layouts.

\end{document}